%% file: main.tex
\documentclass[letterpaper, 10 pt, conference]{ieeeconf} 
\IEEEoverridecommandlockouts                              

\usepackage{times}

\usepackage{multicol}
\usepackage[bookmarks=true]{hyperref}

\usepackage{hyperref}
\usepackage{amssymb}  
\usepackage{algorithm} 
\usepackage{algorithmic}
\usepackage{graphicx}
\graphicspath{ {figs/} }
\usepackage{float}
\usepackage{
    amsmath,
    float,
    xcolor,
    marginnote,
    subcaption,
    pgfgantt,
    times,
    outlines,
    wrapfig,
    romannum,
    csquotes,
    bm,
    upgreek,
}

\hypersetup{%
  colorlinks=true,
  linkcolor=red,
  linkbordercolor=white,
}
\usepackage{lipsum}
\usepackage{soul,xcolor}

\setstcolor{red}
\usepackage{tabularx}
\usepackage{adjustbox}

\usepackage[doi=false,isbn=false,defernumbers=true,style=numeric,,backend=biber,maxbibnames=1000,sorting=none,firstinits=true]{biblatex}

\newcommand{\methodname}{\textit{ODA}}

\newcommand{\bs}{{\mathbf s}}
\newcommand{\ba}{{\mathbf a}}

\DeclareMathOperator*{\argmax}{arg\,max}

\makeatletter
\let\blx@rerun@biber\relax
\makeatother
\begin{document}



\title{Offline Meta-Reinforcement Learning for Industrial Insertion
\thanks{*Equal contribution} 
\thanks{$^{1}$X, The Moonshot Factory, Mountain View, CA, USA}%
\thanks{$^{2}$Intrinsic Innovation LLC, Mountain View, CA, USA}%
\thanks{$^{3}$Deepmind, London, UK}%
\thanks{$^{4}$Google Brain, Mountain View, CA, USA}%
\thanks{$^{5}$Department of Electric Engineering and Computer Science, University of California, Berkeley, Berkeley, CA, USA}
\thanks{$^{6}$Stanford University (work done as an intern at X)}%
}

\author{Tony Z. Zhao$^{*1,6}$, Jianlan Luo$^{*2}$, Oleg Sushkov$^{3}$, Rugile Pevceviciute$^{3}$, \\ Nicolas Heess$^{3}$, Jon Scholz$^{3}$, Stefan Schaal$^{2}$ and Sergey Levine$^{4,5}$ }


\maketitle

\begin{abstract}
Reinforcement learning (RL) can in principle let robots automatically adapt to new tasks, but current RL methods require a large number of trials to accomplish this. 
In this paper, we tackle rapid adaptation to new tasks through the framework of meta-learning, which utilizes past tasks to learn to adapt with a specific focus on industrial insertion tasks.
Fast adaptation is crucial because prohibitively large number of on-robot trials will potentially damage hardware pieces. Additionally, effective adaptation is also feasible in that experience among different insertion applications can be largely leveraged by each other.
In this setting, we address two specific challenges when applying meta-learning. 
First, conventional meta-RL algorithms require lengthy online meta-training. We show that this can be replaced with appropriately chosen offline data, resulting in an offline meta-RL method that only requires demonstrations and trials from each of the prior tasks, without the need to run costly meta-RL procedures online.
Second, meta-RL methods can fail to generalize to new tasks that are too different from those seen at meta-training time, which poses a particular challenge in industrial applications, where high success rates are critical. We address this by combining contextual meta-learning with direct online finetuning: if the new task is similar to those seen in the prior data, then the contextual meta-learner adapts immediately, and if it is too different, it gradually adapts through finetuning. 
We show that our approach is able to quickly adapt to a variety of different insertion tasks, with a success rate of 100\% using only a fraction of the samples needed for learning the tasks from scratch. Experiment videos and details are available at \url{https://sites.google.com/view/offline-metarl-insertion}.

\end{abstract}

\input{sections/01_introduction.tex}\label{introduction}
\input{sections/02_background.tex}\label{problem}
\input{sections/03_related_work.tex}\label{related work}
\input{sections/04_methods.tex}\label{methods}
\input{sections/05_experiments.tex}\label{experiments}
\input{sections/06_analysis}\label{experiments}
\input{sections/07_conclusion.tex}\label{conclusions}

\printbibliography
\end{document}

%% file: sections/01_introduction.tex
\section{INTRODUCTION}\label{sec:intro}

Industrial robotic manipulation tasks are gaining more traction than ever, partially due to the sharply increasing interest in AI-enabled applications and more suitable and affordable robot hardware \cite{yahoorobotics,intrinsic_blog,intrinsic_reuters}. 
Part mating and part insertion are very common in the manufacturing industry, and the goal to enable high-mix low-volume scenarios are among the most prominent research topics \cite{wang_future_manufacturing}.
Current manufacturing methods often rely on human-engineered heuristics that must be finetuned or redesigned for each new instance.
In this context, reinforcement learning provides an appealing and automated alternative.
Previous works have demonstrated that reinforcement learning can solve some of these manipulations tasks \cite{vecerik2017leveraging, Luo-RSS-21, vecerik2019practical, luo2019reinforcement, visual_residual_rl,levine2016end,LevineWA15, tuomas2018manipulation,InoueHighPrecision, aviv2017icra, Vecerk2017LeveragingDF, Thomas2018LearningRA}. 
However, lengthy exploration processes are often required. For example, it would usually take a RL algorithm hours to solve one connector insertion task \cite{Luo-RSS-21}. This process needs to be repeated for each individual connector, even though these connectors can bear similar insertion strategies; e.g., the knowledge to insert a USB-A connector should provide guidance for inserting a USB-C connector.



\begin{figure}[t]
\vspace{0.10cm}
\centering
\includegraphics[clip,width=1.0\columnwidth]{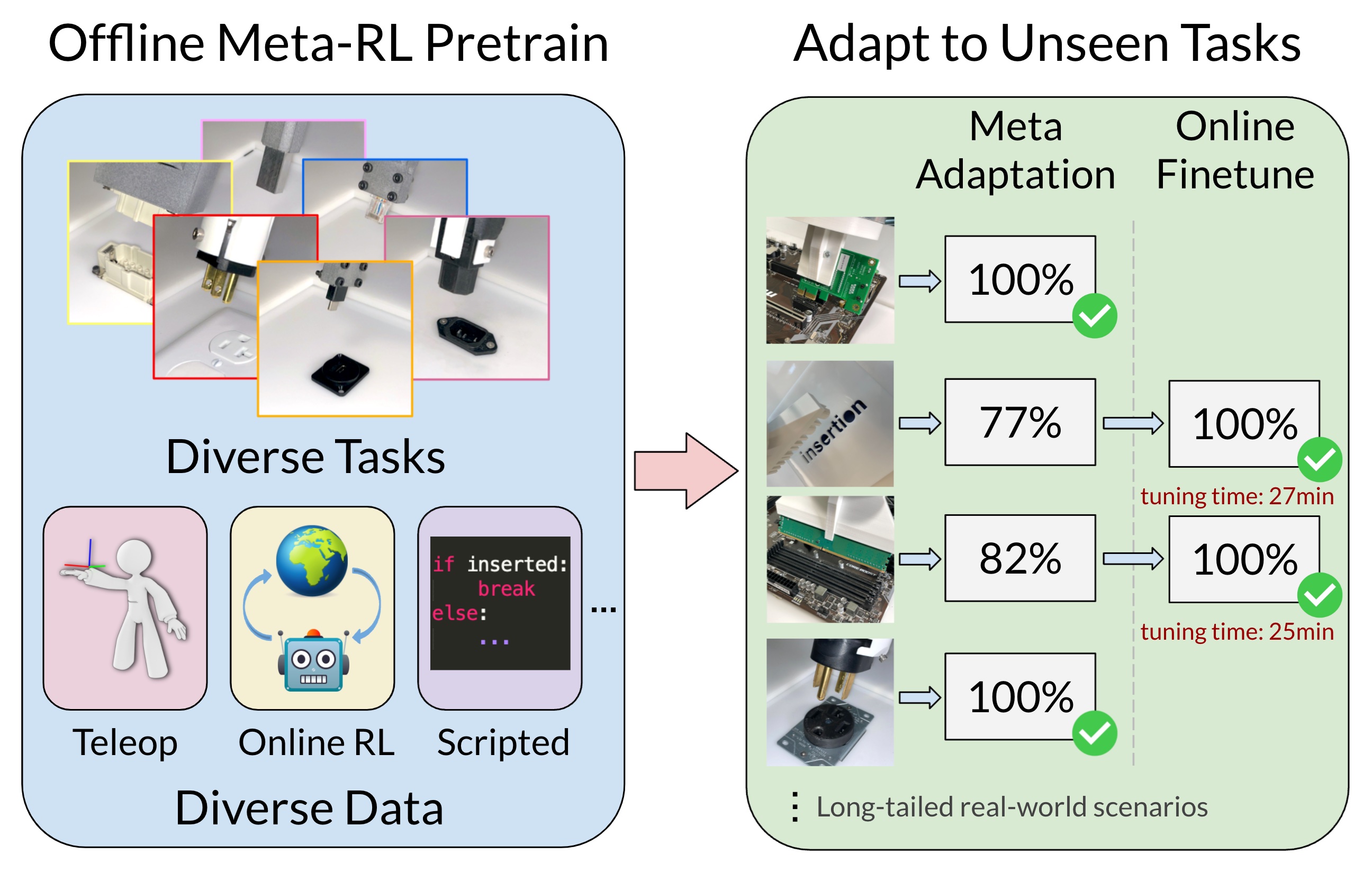}
\caption{Our approach, \textit{Offline Meta-RL with Demonstration Adaptation} (\methodname), uses offline data for pretraining, then adapts using both demonstrations and online finetuning.}
\vspace{-0.5cm}
\label{fig:teaser}
\end{figure}

\begin{figure*}[tbh]
\centering
\includegraphics[clip, width=0.97\textwidth]{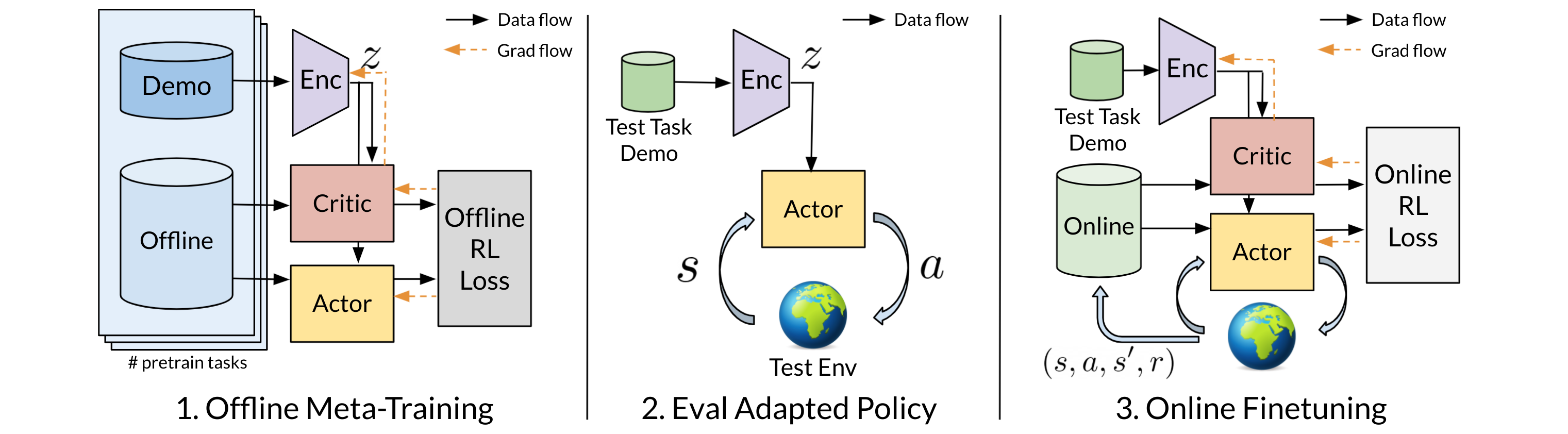}
\caption{During offline meta-training (left), \methodname{} trains the policy using offline data for each training task, with an encoder that learns to summarize the demonstrations into a latent variable $z$.
Next, during adaptation to a new task (middle), we condition the encoder on demos for the new task, obtaining a policy that either solves the new task immediately, or else provides a initialization for the third stage (right), where the entire policy is finetuned end-to-end with online interactions.}
\vspace{-0.5cm}
\label{fig:algo_diag}
\end{figure*}


Meta-RL algorithms can make it possible to adapt to new tasks quickly by pre-training on a distribution of tasks \cite{finn-maml,pearl,Gupta2018MetaReinforcementLO,duan2016rl2,Anusha2018meta,Wortsman2019LearningTL, meld, gerrit2020metarl}. 
However, the lengthy online meta-training phase required can be prohibitive in industrial robotics settings. 
On the other hand, robots that are already deployed at manufacturing sites can generate large amounts of useful data from their existing controllers. This makes it appealing to consider incorporating ideas from offline RL \cite{levine2020offline, kumar2020conservative, nair2020accelerating, singh2020cog, Dorfman2020OfflineML, Mandlekar2020IRISIR, Fujimoto2021AMA}, which utilizes previously collected datasets in place of active data collection. 
Several works have proposed offline algorithms for meta-RL \cite{Dorfman2020OfflineMR, pong2021offline, Mitchell2021OfflineML, aviv2021offline, Li2021EfficientFM}. 
Our offline meta-RL approach builds on Pong et al.~\cite{pong2021offline}, but incorporates a number of design choices that make it practical for industrial applications: (i) we leverage demonstrations for both training tasks and the test task, which reduces the need for lengthy exploration and accelerates adaptation; (ii) we combine meta-learned adaptation via a PEARL-style encoder~\cite{pearl} with full finetuning of the entire policy, such that even when the new task is very different from those seen in meta-training, the policy can still adapt gradually until it succeed. 

Concretely, our method first meta-trains a policy on available datasets. For a new task, we adapt the policy using user-provided demonstrations. This phase may already solve most tasks, but if it does not, we follow up with a fine-tuning phase until satisfactory performance is attained.
The offline training dataset can comprise demonstration data from humans, replay buffer data from RL training, or even data from hand-designed (``scripted'') policies. During meta-training, we first pass human demonstration data into an encoder to infer the task variable $z$; we then condition both the actor and critic on $z$ in addition to observations or actions. All networks are trained jointly with offline RL. This encourages the encoder to pick up task information from demonstrations that is useful for actor and critic. When adapting to a new task, we infer a new task variable from a few human demonstrations and evaluate the adapted policy. We only finetune the policy end-to-end with additional online samples when adaptation alone does not solve the task.

Our contributions are summarized as follows: We present a novel offline meta-RL framework that integrates a number of design choices to make it practical in realistic industrial insertion settings. 
Our method can adapt to unseen tasks reliably and rapidly by leveraging static offline datasets and optional on-policy robot interaction. 
We evaluate it with a total of 13 industrial insertion tasks, including ones like RAM insertion that has tight tolerance and requires delicate handling. Our method is able to achieve 100/100 success for all 13 tasks within 30 minutes of adaptation.


%% file: sections/02_background.tex
\section{BACKGROUND}\label{sec:background}
In this section, we give a brief background of RL and offline meta-RL. We consider a Markov Decision Process (MDP) associated with task $\mathcal{T}$ defined by $\mathcal{M}_{\mathcal{T}} = \{\mathcal{S}, \mathcal{A}, \mathcal{P}, \mathcal{R}, p_0, \gamma \}$. $\mathcal{S}$ is the state space, $\mathcal{A}$ is the action space, $\mathcal{P}$ governs the transition dynamics, $\mathcal{R}$ is the reward function, $\gamma$ is the discount factor, $p_0(s)$ is the initial state distribution. The replay buffer for the task $\mathcal{D}_{\mathcal{T}}$ consists of state, action, reward, next state tuples $(s, a, r, s^{\prime})$. We also define the associated Q function of the policy as \mbox{$Q^{\pi}(s_t, a_t) = \sum_{t^{\prime}=t}^{T} \mathbb{E}_{\pi}[\gamma^{t^{\prime}-t}r(s_{t^{\prime}}, a_{t^{\prime}}) |  s_t, a_t] $}, with $a_{t^{\prime}}$ taken with the current policy $\pi$.
We can learn this Q-function via Bellman backups in actor-critic methods \cite{Lillicrap2016ContinuousCW}, and the objective of reinforcement learning is to find the policy that maximize cumulative discounted rewards $R=\sum_{t}  \gamma^{t} r(s_{t},a_{t})$.

In the offline meta-RL setting, we assume a task distribution $p_{\mathcal{T}}(\mathord{\cdot})$, and every task $\mathcal{T}$ is an aforementioned MDP. We slightly abuse the notion of $\mathcal{D}_{\mathcal{T}}$ to represent the static dataset linked to each task.
During offline meta-training, we have N training tasks $\mathcal{T}_{i=1, ...,N}$ sampled from $p_{\mathcal{T}}(\mathord{\cdot})$ and we assume access to dataset $\mathcal{D}_{\mathcal{T}_i}$ for each sampled task.
We optimized the appropriate meta-training loss using the datasets to learn a meta-policy $\pi_{\theta^{\star}}$ and a meta-adaptation procedure $f_{\phi^{\star}}$. 
At meta-test time, we face a new task $\mathcal{T'}$ sampled from $p_{\mathcal{T}}(\mathord{\cdot})$, and we assume access to the associated new dataset $\mathcal{D}_{\mathcal{T'}}$.
The meta-adaptation procedure adapts the meta-policy to the new task, i.e. $\hat{\pi} = f_{\phi^{\star}}(\pi_{\theta^{\star}}, \mathcal{D}_{\mathcal{T'}})$, and we aim to maximizes the return of $\hat{\pi}$:
\begin{align}
    \theta^{\star}, \phi^{\star} = \text{argmax}_{\theta, \phi} \,\,  \sum_{t=0}^{T} [\gamma^{t}r(s_{t}, a_{t})];
\end{align}
where $s_{t}, a_{t}$ denote the state and action taken at step $t$ when executing the adapted policy $\hat{\pi}$ in $\mathcal{T'}$.

%% file: sections/03_related_work.tex
\section{RELATED WORK}\label{sec:related_work}

In addition the works cited in Sec.~\ref{sec:intro}, we discuss several previous works that focus on insertion problems.
Heuristic-based methods combined with compliance-controlled robotic arms or even passive mechanisms \cite{Haskiya1999RoboticAC, Bruyninckx1995PegonholeAM, Chhatpar2001SearchSF, Whitney1977ForceFC, whiteneypassive, Park2013IntuitivePA, intelligentinsertion, compliant_peg_hole, tang2015learningpeg} can solve some insertion tasks reliably, but such methods typically require manual tuning for every type of task and connector. Gerrit et al.~\cite{gerrit2020metarl} applied PEARL\cite{pearl} for solving industrial insertion tasks in a sim2real fashion, while it has yet to be shown that such domain randomization approach could scale beyond the two insertion tasks considered in the paper. 
We posit that it would require a significantly larger family of parameters to randomize over in order to capture the complex physics like contact dynamics and deformation, making it computationally expensive and involves careful engineering.
While our paper show that a broader range of connectors insertion problems can be solved by collecting data in the real world and process them with offline meta-RL.
Another relevant work is by Spector et al.~\cite{Spector2021InsertionNetA}, which learns a residual insertion policy on top of an existing PD controller through ``backward learning." Unlike \cite{residualrl,visual_residual_rl}, the residual policy is learned in a supervised fashion for sample efficiency. They generate the training data by a scripted policy producing random perturbations starting from the goal, and show this type of template policy can in principle tackle a series of industrial insertion tasks. However, as the authors pointed out, this method has trouble inserting connectors with small clearance, thus resulting in lower performance for connectors like USB-C, whereas our method is able to consistently adapt to 13 new connectors with a 100\% success rate, as shown in our experiments. The paper also focused on visual adaptation, which can be a different challenge than ours.

%% file: sections/04_methods.tex
\section{Offline RL with Demonstration Adaptation}





In our method, which is summarized in Fig.~\ref{fig:algo_diag}, an adaptive policy is meta-trained via offline RL on a user-provided dataset consisting of trials for a variety of different tasks, together with corresponding policies for each task. This adaptive policy is then used to solve new tasks, first by adapting to them using a small number of demonstrations, and then by finetuning with online RL.
In the meta-training stage, we assume access to two buffers containing demonstrations and offline data respectively, for each of the pretrain tasks. Our goal is to meta-train the policy so that it can adapt to new tasks using user-provided demonstrations, which are easy and safe to provide for a new task in limited quantity, rather than adapting entirely via online exploration trials as in prior work. Our meta-learned policy consists of a policy network $\pi_\theta(a|s,z)$ that is conditioned on the current state $s$ and a \emph{latent code} $z$ that denotes the task, as well as an encoder network $q_{\phi}(z | c)$ that predicts $z$ for each task from a \emph{context} $c$, which consists of the demonstrations. Intuitively, this encoder extracts the information necessary to determine how to perform the task from the demos and consolidates it into $z$, as we will discuss in this section. This design resembles prior methods such as PEARL and MELD~\cite{pearl, meld}, with the main difference that we specifically encode demonstrations rather than online experience. However, unlike meta-imitation methods~\cite{finn2017one}, the policy and encoder are still meta-trained with RL.

When our method needs to adapt to a new task (e.g., insert a new type of connector), we assume that we are first provided with demonstrations, which form the context $c$ for the new task and which we use to infer $z \sim q(z | c)$. Then, we can run $\pi(a|s,z)$. In many cases, this policy will already perform the task successfully. However, if the new task is substantially different from the tasks seen during meta-training, the encoder may be unable to infer a $z$ for which the policy is consistently successful. In this case, we can finetune the entire policy and encoder with reinforcement learning and additional online data collection. Typically, this finetuning process is quite fast (5-10 minutes in our experiments), since $\pi(a|s,z)$ already provides a very good initialization. We describe the components of our method in more detail below.


\subsection{Contextual Meta-Learning with Demonstrations}

The design of our meta-reinforcement learning model resembles prior contextual meta-RL algorithms, such as PEARL and MELD~\cite{pearl, meld}, with the principle differences being that we utilize offline data and adapt using demonstrations. In this design, adaptation is performed via an encoder network $q_{\phi}(z | c)$, which reads in a context $c$ consisting of some trajectories for the current task, and infers a vector of latent variables $z$ that is passed to the policy $\pi_\theta(a|s,z)$. In this way, $z$ encapsulates the information needed to perform the current task. We follow prior work and use the same architecture for $q_{\phi}(z | c)$ as PEARL.
Training is performed end-to-end via RL, which trains both $\pi_\theta(a|s,z)$ and $q_{\phi}(z | c)$ together to maximize performance on the meta-training tasks. We discuss the specific algorithm we use in Section~\ref{sec:rl}. We deviate from prior work in our choice of context $c$: while PEARL and MELD~\cite{pearl, meld} use trials from the latest policy to generate trajectories for the context, our method uses user-provided demonstrations. Such demonstrations are typically easy to provide in small quantities, and spare the method from needing to explore via trial and error. In our experiments, we collect demonstrations using a space mouse to control the tool-center-point (TCP) pose. To meta-train $q_{\phi}(z | c)$ to utilize demonstrations, we maintain two buffers during meta-training: a demonstration buffer, which is used to form the context for each minibatch, and another buffer that consists of all offline data and is used to sample the transitions for actor-critic updates. Following Rakelly et al.~\cite{pearl}, we add a KL-divergence loss to $q_{\phi}(z | c)$ to minimize unnecessary information content in $z$. This is summarized in Algorithm~\ref{alg:meta-train}.


In the offline setting, demonstrations also allow us to sidestep the train-test distribution shift problem in offline meta-RL pointed out by Pong et al.~\cite{pong2021offline}. When using the meta-trained policy to collect data to form the context $c$, the data collected for a new task will differ systematically from the offline data used for training $q_{\phi}(z | c)$ during the offline meta-training phase, since the learned policy differs from the policy that collected the offline data, and as pointed out by Pong et al.~\cite{pong2021offline}, this can lead to very poor performance for standard offline meta-RL methods. Since we instead use demonstrations, which do not experience this distributional shift, our method performs well even with purely offline meta-training.

\subsection{Offline and Online Reinforcement Learning}
\label{sec:rl}

In order to meta-train on offline data, we require a suitable offline RL algorithm. Since our aim is to meta-train a policy that can then be fine-tuned to new tasks, we use Advantage-Weighted Actor-Critic (AWAC)~\cite{nair2020accelerating}, which has been shown to perform well in offline training followed by fine-tuning. We summarize AWAC in this section. AWAC aims to learn a policy that maximizes reward while bounding the deviation from the data distribution $\pi_{\beta}$:
\begin{align*}
    \theta^{\star} =
    \argmax_{\theta} \mathbb{E}_{\bs \sim \mathcal{D}}\mathbb{E}_{\pi_{\theta}(\ba | \bs)}[Q_{\phi}(\bs,\ba)] \, \mathbf{s}.\mathbf{t}. \, D_{KL}(\pi_{\theta} \| \pi_{\beta}) \leq \epsilon 
\end{align*}
This constrained optimization can be solved via Lagrangian duality, and the solution can be approximated via weighted maximum likelihood:
\begin{align*}
    \theta^{\star} = \argmax_{\theta} \mathbb{E}_{\bs, \ba \sim \beta}[\log\pi_{\theta}(\ba | \bs) \exp(A^{\pi}(\bs,\ba))], 
\end{align*}
where $A^{\pi}(\bs,\ba) = Q_\phi(\bs,\ba) - E_{\ba \sim \pi_\theta(\ba|\bs)}[Q_\phi(\bs,\ba)]$ is the advantage under current policy $\pi$.  
The critic $Q_\phi$ is trained to minimize the standard Bellman error.
For further detail, we refer to prior work~\cite{nair2020accelerating}. 


This RL algorithm is used to meta-train the policy, critic and encoder, and also to finetune it in the last stage of adaptation. The design of our approach resembles the method of Pong et al.~\cite{pong2021offline}, with several critical differences: (i) We condition the encoder on demonstrations, making it possible to adapt to new tasks without risky exploration. This is important in safety-critical industrial settings. (ii) Unlike Pong et al.~\cite{pong2021offline}, we do not require any online meta-training phase, because the conditioning demonstrations do not incur distributional shift. This makes our method significantly easier to meta-train, since all meta-training uses offline data with no online collection at all. (iii) We utilize a hybrid adaptation procedure where we \emph{first} use the encoder to adapt to demonstrations, and then finetune further with online AWAC to maximize performance. As we show in our experiments, this finetuning stage can significantly improve performance on difficult tasks. This combination of design decisions is novel to our approach. Though it does combine a number of parts that are based on prior work, as we demonstrate in our experiments, the particular combination of parts we employ is important for good final performance. We describe both stages in detail in Algorithm \ref{alg:meta-train} and \ref{alg:meta-test}. To avoid clutter, we describe the algorithm when batch size is 1 during data sampling (e.g. Algo.~\ref{alg:meta-train} line 4, 5), when in practice we sample a mini-batch. For reproducibility, we include network architectures, hyperparameters and other implementation details on our \href{https://sites.google.com/view/offline-metarl-insertion}{website}.

\begin{algorithm}[t]
\caption{\bf \methodname{} Meta-training}
\begin{algorithmic}[1]
\REQUIRE $D^{i}_{demo}$ and $D^{i}_{offline}$ for each of $N$ training tasks, learning rates $\eta_1, \eta_2, \eta_3$, temperature $\lambda$, KL weight $\beta$
\STATE Init. encoder $q_{\phi}$, actor $\pi_{\theta}$, critic $Q_{\psi}$
\WHILE{not converged}

\FOR{task $i=1, 2, ..., N$}
\STATE Sample demo data as context $c_{i} \sim D^{i}_{demo}$
\STATE Sample offline data $(s, a, s', a', r) \sim D^{i}_{offline}$ 

\STATE Sample task variable $z_{i} \sim q_{\phi}(\cdot |c_{i})$ 
\STATE $y = r(s, a) + \gamma \mathop{\mathbb{E}}_{s', a' \sim D^{i}_{offline}}Q_{\psi}(s', a', z_{i}) $
\STATE $\mathcal{L}^i_{critic} = (Q_{\psi}(s, a, z_{i}) - y)^{2}$
\STATE $\mathcal{L}^i_{actor} = -\log \pi_{\theta}(a | s, z_{i}) \exp(\frac{1}{\lambda} A^{\pi}(s, a, z_i))$
\STATE $\mathcal{L}^i_{KL} = \beta D_{KL}(q_{\phi}(\cdot | c_{i}) \| \mathcal{N}(0, I)) $
\ENDFOR



\STATE $\phi \gets \phi - \eta_1 \nabla_\phi \sum_i (\mathcal{L}^i_{critic} + \mathcal{L}^i_{KL}) $
\STATE $\theta \gets \theta - \eta_2 \nabla_\theta \sum_i \mathcal{L}^i_{actor} $
\STATE $\psi \gets \psi - \eta_3 \nabla_\psi \sum_i \mathcal{L}^i_{critic} $

\ENDWHILE

\end{algorithmic}
\label{alg:meta-train}
\end{algorithm}



\begin{algorithm}[tbph]
\caption{\bf \methodname{} Adaptation and Finetuning}
\begin{algorithmic}[1]
\REQUIRE Test task demo $D_{test}$, learning rates $\eta_1, \eta_2, \eta_3$, temperature $\lambda$, KL weight $\beta$, Pretrained $\pi_{\theta}$, $Q_{\psi}$, $q_{\phi}$
\STATE Init. empty online buffer $\mathcal{B}$

\STATE Sample demo data as context $c \sim D_{test}$ 
\STATE Sample task variable $z \sim q_{\phi}(\cdot | c)$

\STATE Evaluate policy $\pi_{\theta}(\cdot | s, z)$, \textbf{exit} if policy solves the task

\WHILE{not converged}
\STATE collect trajectory $\uptau$ by executing policy $\pi_{\theta}(\cdot | s, z)$
\STATE add $\uptau$ to $\mathcal{B}$

\STATE Sample demo data as context $c \sim D_{test}$ 
\STATE Sample offline data $(s, a, s', r) \sim \mathcal{B}$ 
\STATE Sample task variable $z \sim q_{\phi}(\cdot | c)$ 
\STATE Calculate $\mathcal{L}_{actor}$, $\mathcal{L}_{critic}$, $\mathcal{L}_{KL}$ same as Algo.\ref{alg:meta-train} 

\STATE $\phi \gets \phi - \eta_1 \nabla_\phi (\mathcal{L}_{critic} + \mathcal{L}_{KL}) $
\STATE $\theta \gets \theta - \eta_2 \nabla_\theta \mathcal{L}_{actor} $
\STATE $\psi \gets \psi - \eta_3 \nabla_\psi \mathcal{L}_{critic} $
\ENDWHILE

\end{algorithmic}
\label{alg:meta-test}
\end{algorithm}

%% file: sections/05_experiments.tex
\begin{figure*}[t]
\captionsetup[subfigure]{labelformat=empty}
\begin{subfigure}{1.0\textwidth}
\centering
\includegraphics[clip,width=1.0\textwidth]{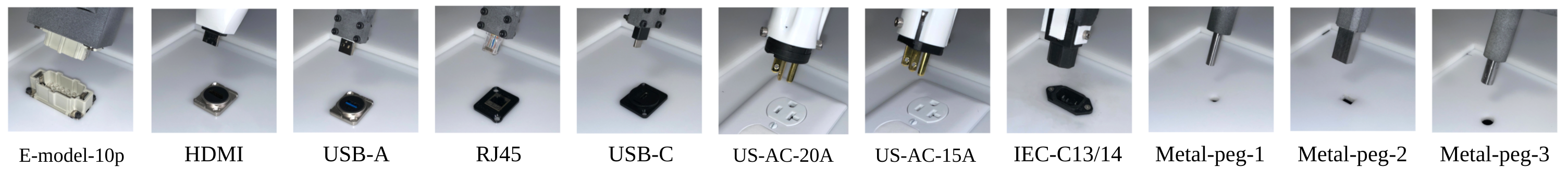}
\end{subfigure}%
\newline
\begin{subfigure}{1.0\textwidth}
\centering
\end{subfigure}%
\vspace{-0.20cm}
\caption{\textbf{Offline training tasks.} For each of the 11 training tasks, we run DDPGfD \cite{vecerik2017leveraging} to collect data for offline training.}
\label{fig:training_tasks}
\end{figure*}

\begin{figure*}[t]
\vspace{-0.30cm} 
\captionsetup[subfigure]{labelformat=empty}
\begin{subfigure}{1.0\textwidth}
\centering
\includegraphics[clip,width=1.0\textwidth]{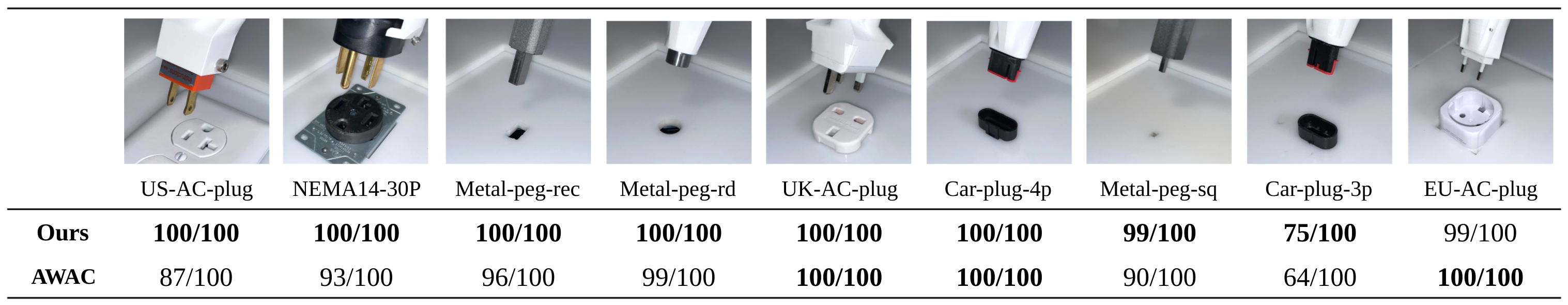}
\end{subfigure}%
\newline
\begin{subfigure}{1.0\textwidth}
\centering
\end{subfigure}%
\caption{\textbf{Test tasks and adaptation performance.} We evaluate adaptation with 9 test tasks and compare it to AWAC \cite{nair2020accelerating}. Our method \methodname{} achieves 100\% success rate for 6/9 tasks, and outperforms AWAC for 8/9 tasks. Adaptation only uses the demos for each task, without additional online interaction.}
\label{fig:pretrain_results}
\end{figure*}

\begin{figure*}[t]
\vspace{-0.50cm} 
\captionsetup[subfigure]{labelformat=empty}
\begin{subfigure}{0.33\textwidth}
\centering
\includegraphics[width=1.0\textwidth]{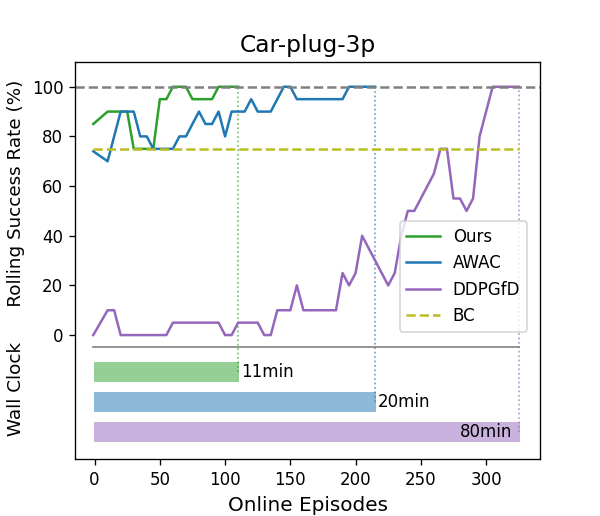}
\end{subfigure}%
\begin{subfigure}{0.33\textwidth}
\centering
\includegraphics[width=1.0\textwidth]{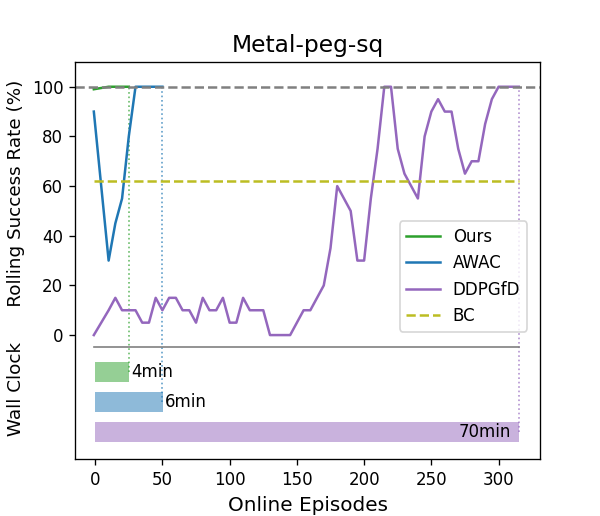}
\end{subfigure}%
\hfill
\begin{subfigure}{0.33\textwidth}
\centering
\includegraphics[width=1.0\textwidth]{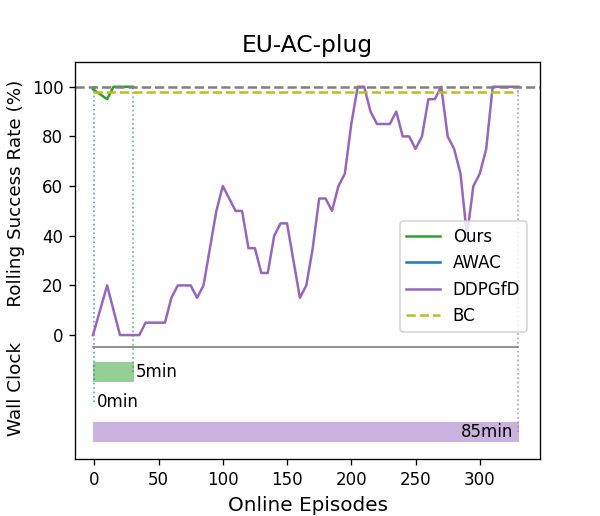}
\end{subfigure}%
\caption{\textbf{Finetuning performance.} For tasks where adapting with demos does not reach 100/100, we finetune with online interaction. We plot the training success rate and wall-clock time for \methodname{}, finetuning AWAC\cite{nair2020accelerating}, DDPGfD\cite{vecerik2017leveraging}, as well as behavioral cloning. Our method reduces wall-clock time by 11.75x compared to DDPGfD, and outperforms AWAC. 
We note that the length of bar plot reflects the number of episodes, and successful episodes are significantly shorter wall-clock.}
\vspace{-0.3cm}

\label{fig:curves}
\end{figure*}


\section{EXPERIMENTS}

We evaluate our algorithm by adapting to 12 new tasks (see Fig.~\ref{fig:pretrain_results} and \ref{fig:challenge_tasks}), studying the following questions:
\begin{enumerate}
    \item How well does our method perform compared to standard offline RL?
    \item Can finetuning provide fast adaptation even for very different novel tasks, where just running the meta-trained encoder fails?
    \item Can our method handle the challenging tasks in Fig.~\ref{fig:challenge_tasks}, which are outside of the training distribution?
    \item Does our meta-adaptation improve with more offline training data?
\end{enumerate}

{\setlength{\parindent}{0cm}
\paragraph{Experiment setup}

In our insertion tasks, the robot starts holding an object (e.g., a connector), and must position it into a goal pose where it is inserted into a socket. The reward is a binary indicator for whether the insertion is successful, and the episode terminates either when the robot succeeds, or the episode length exceeds a maximum time limit.
We run experiments with a KUKA iiwa7 robot. The agent controls the TCP twist of the robot at 10Hz, which is tracked and interpolated by a downstream impedance controller at 1000 Hz. The observation to our agent consists of the robot's TCP pose, velocity, and the wrench measured at the tool tip.  
For both training and test tasks, we express robot TCP  information in a relative coordinate system, with its origin located at the TCP, as in prior work~\cite{Luo-RSS-21}. We add perturbation noise from a uniform distribution $\mathcal{U}$[-1mm, 1mm] at the beginning of each episode, at both training and test time. The policy does not have access to this noise, and therefore must be robust to the perturbations.

\paragraph{Offline dataset}
We collected data for 11 different plug-socket pairs, as shown in Figure~\ref{fig:training_tasks}. The data was collected by running DDPGfD \cite{vecerik2017leveraging} and saving the entire replay buffer as offline data for our method, as well as demonstrations for each task.
On average, we obtain 500 episodes of RL interaction data and 20 episodes of human demonstration data for each of the training tasks.
To further improve the diversity of the dataset, we also reran the final DDPGfD policies with added noise to provide good coverage of the workspace. We list dataset details in our \href{https://sites.google.com/view/offline-metarl-insertion}{website}.
}


\subsection{Adaptation via Learned Encoder}

In our first set of experiments, we evaluate how well our learned encoder can adapt to new tasks using only the demonstrations for those tasks. We use 9 test tasks as shown in Fig.~\ref{fig:pretrain_results}. We collect 20 demonstrations for each task and feed them into our encoder to infer the task information $z$. We compare to standard AWAC pretrained on the same data, with demos added into the offline data.
As shown in Fig.~\ref{fig:pretrain_results}, our method performs well across all of the tasks, succeeding 100/100 times on 6 of the 9, while AWAC attains comparatively lower success rates.

AWAC learns a single policy for all connectors, and relies on this policy to generalize effectively, whereas our approach can adapt to each task using a small amount of demonstration data.
Let's look at a motivating example to further illustrate this difference. We tested pretrained policies both from our method and AWAC on one of the training tasks, ``E-Model-10p" (see Fig.~\ref{fig:training_tasks}). Since this is a training task, we expect both methods to do well, but AWAC only has a success rate of 49\%, while our method gets 100\%. This task requires fine-grained adjustments to insert, which are quite distinct from other connectors in the dataset, making it hard for the non-adaptive AWAC policy to succeed on this task and all the others with the same strategy. 
On the other hand, our method succeeds by determining the right strategy from the provided demonstrations. Of course, we could finetune AWAC with additional data, which would make for a more fair comparison. We study precisely this in the next set of experiments.

\subsection{Handling Out-of-Distribution Tasks via Finetuning}
\label{finetune_section}

If the task at test-time is out-of-distribution relative to tasks seen during offline training, simply utilizing the demos with the encoder might not be enough to succeed. In this section, we study how finetuning can alleviate this issue, using the three tasks where our method fails to achieve 100\% performance from only the demos. In Fig.~\ref{fig:curves}, we show finetuning results for (1) our method, (2) finetuning AWAC with demonstrations added to initial buffer, (3) training DDPG from demos,
and (4) the performance of behavioral cloning with the demos. 
(4) fails to reach 100\% because imitating human demonstrations will not produce a policy that is robust to starting pose randomization.
(1), (2), and (3) all attain a success rate of 100/100 after finetuning, while our method finetunes significantly faster as shown in the bar plot below each graph in Fig.~\ref{fig:curves}. On average, we finish training 11.75 times faster than DDPGfD. 
This is not completely surprising because the power of offline training: it allows us to leverage the rich information in offline datasets. 
We also find that in the two tasks where our method outperforms AWAC (left and middle plot of Fig.~\ref{fig:curves}), we finish training 1.73 times faster than AWAC. For the third task (right plot of Fig.~\ref{fig:curves}), our algorithm takes 5 mins to account for the performance difference. Note the speedup comparison here is conservative: we do not include the tasks where AWAC fails but our method achieves 100\% without any finetuning at all (i.e., the first 4 tasks in Fig.~\ref{fig:pretrain_results}).
It's also worth noting that we have tried to pretrain a policy with DDPG on the offline datasets, but it showed 0\% success rate on test tasks. We then finetune the policy pretrained with DDPG loss, while it takes similar amount of time to succeed as randomly initialized networks. This validates the necessity of offline RL training. We did not include it in the figure for presentation clarity. 


\subsection{Out-Of-Distribution Challenge Tasks}
\begin{figure}[!h]
\centering
\includegraphics[clip,width=0.9\columnwidth]{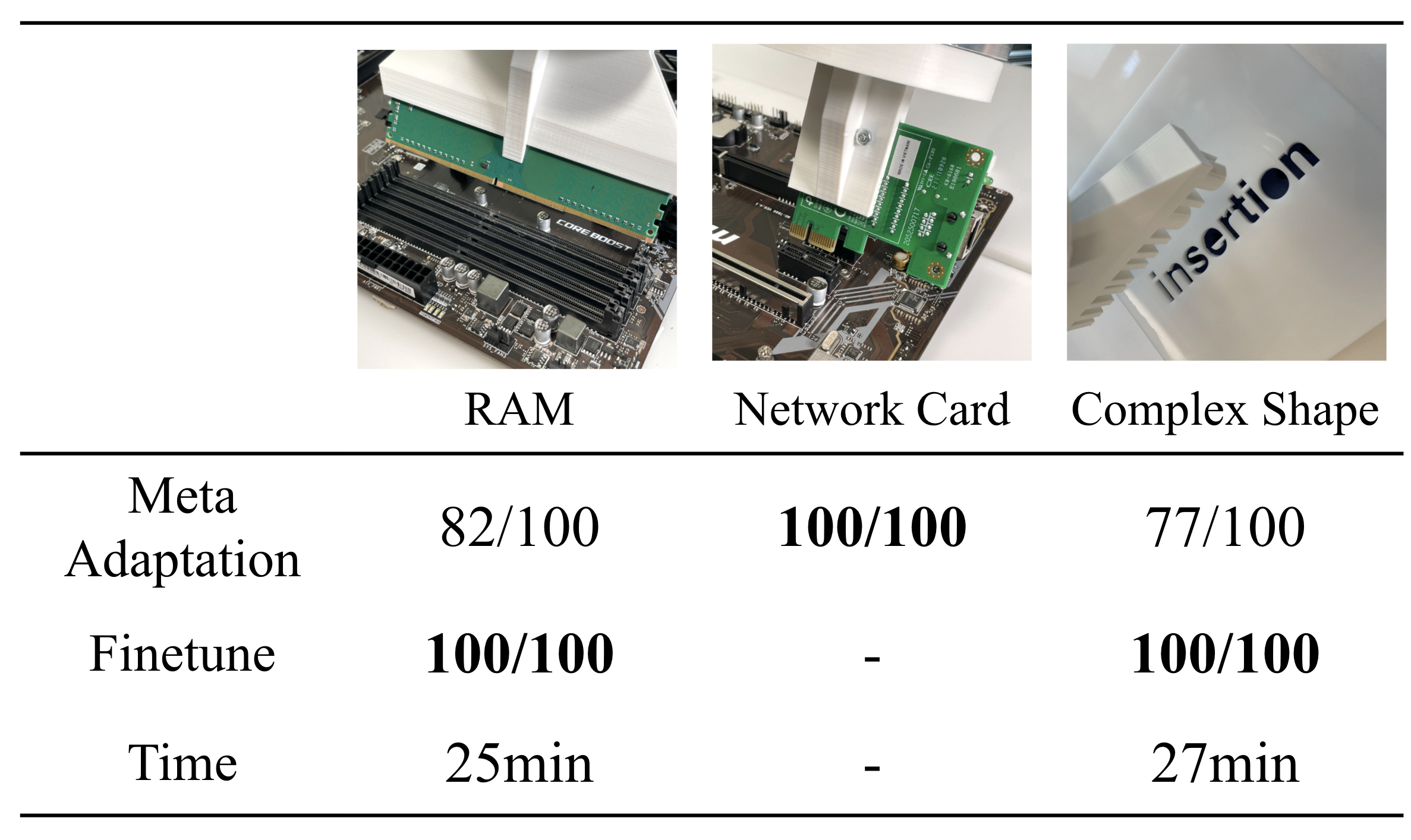}
\caption{\textbf{Challenge tasks} We challenge our method with three additional out of training distribution tasks. \methodname{} is able to solve all three with 100/100 success within 30 minutes.} 
\vspace{-0.1cm}
\label{fig:challenge_tasks}
\end{figure}

In realistic industrial assembly tasks, the robot may be tasked with problems that differ significantly from the conditions seen in training. To study how our method performs in these settings, we evaluate three additional challenge tasks that differ significantly from the training connectors and present a particular physical challenge (see Fig.~\ref{fig:challenge_tasks}): (1) RAM insertion, (2) network card insertion, and (3) complex shape insertion. (1) and (2) are require handling delicate electronics and significant application of force. Training from scratch would be impractical here, since an untrained policy would easily damage the circuit board. (3) is challenging because of the small clearance and the very different contact shape compared to training tasks. We do \textbf{not} inject uniform noise to the starting pose to make these three tasks more tractable. 

As shown in Figure~\ref{fig:challenge_tasks}, our method is able to solve all three tasks 100/100 within 30 minutes. For network card insertion, we are able to succeed without any finetuning. We posit that inserting the RAM is significantly harder than the network card because it is much longer (288 vs. 18 pins), requires high precision along the yaw axis, and requires considerable application of force. These characteristics makes it very different from all the training tasks, and generally difficult for a robot to perform. During the 25 minutes required to adapt the policy to perform this task 100\% of the time, no damage is caused to the RAM or the motherboard.

\subsection{Scaling with More Data}

\begin{figure}[!h]
\vspace{-0.40cm}
\centering
\includegraphics[clip,width=0.7\columnwidth]{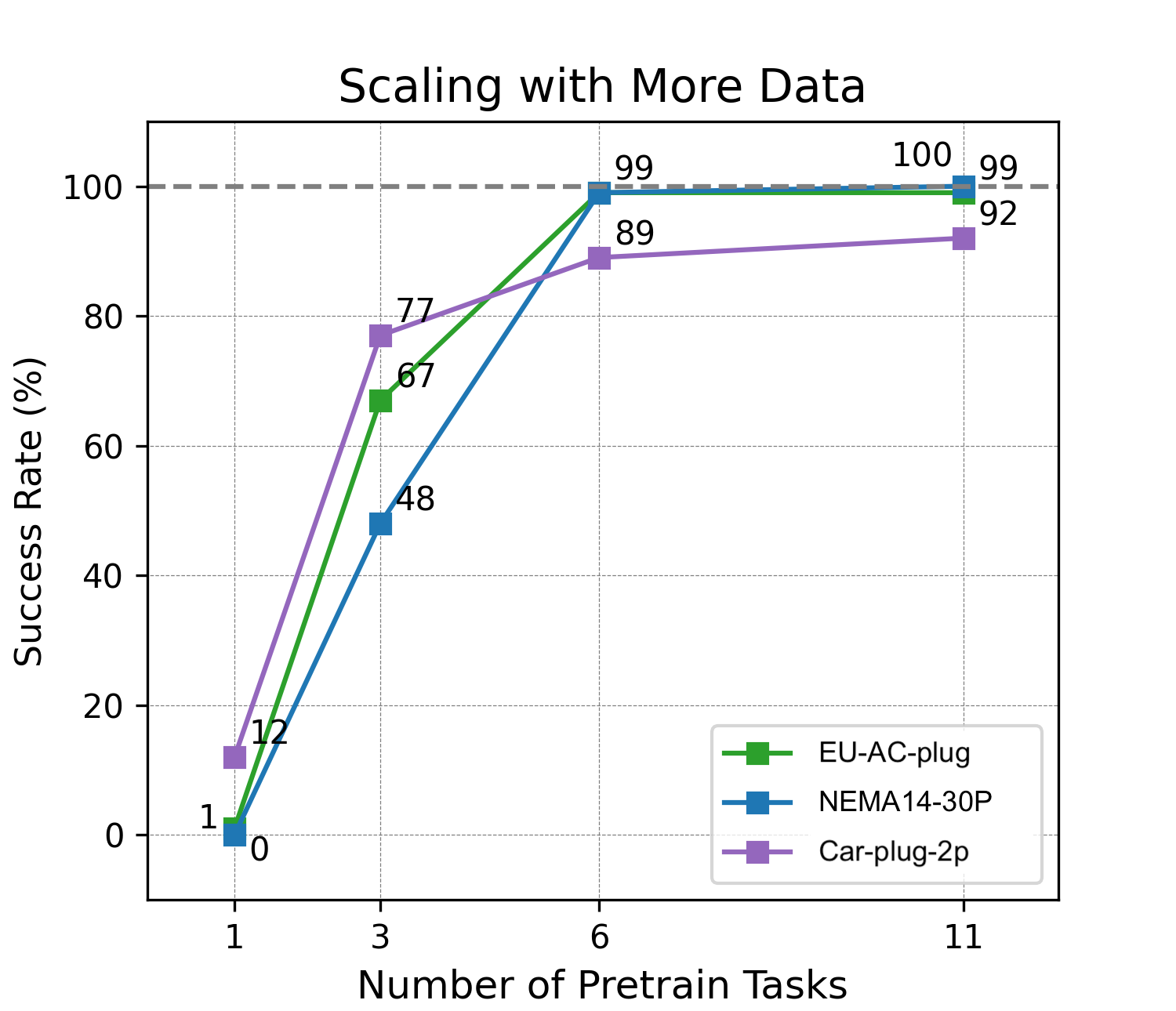}
\vspace{-0.1cm}
\caption{\textbf{Scaling with more data.} We plot the adaptation performance of three tasks as the number of training tasks increases. The success rate increases steadily as the algorithm has access to more data.}
\vspace{-0.1cm}
\label{fig:scaling}
\end{figure}

Lastly, we analyze how the number of training tasks affects adaptation performance of our method, shown in Fig.~\ref{fig:scaling}.
We evaluate the meta-adaptation performance across three tasks, and it continues to improve as we increase the number of tasks in the dataset from 1 to 11.






%% file: sections/06_analysis.tex





%% file: sections/07_conclusion.tex
\section{DISCUSSION}

We introduced an offline meta-RL algorithm, ODA, that can meta-learn an adaptive policy from offline data, quickly adapt based on a small number of user-provided demonstrations for a new task, and then further adapt through online finetuning. We show that this approach is highly effective at adapting to new connector insertion tasks, learning to insert many connectors with a 100\% success rate after just the initial demonstrations, and finetuning other connectors to a 100\% success rate with minutes of additional online training. Although our algorithm is composed of parts proposed in prior work, the particular combination and the use of demonstrations for adapting in contextual meta-RL is novel to our method. Promising future directions include incorporating visual perception, extending the approach to other tasks, and extending it into a lifelong learning framework where each new task is included into a lifelong meta-training process.
